\documentclass{article}

\usepackage{multicol}
\usepackage{amsmath} 
\usepackage{mathtools}
\usepackage{csquotes}
\usepackage{array}
\usepackage{booktabs}
\usepackage{makecell} 
\usepackage[table]{xcolor}

\usepackage[utf8]{inputenc} % allow utf-8 input
\usepackage[T1]{fontenc}    % use 8-bit T1 fonts
\usepackage{url}            % simple URL typesetting
\usepackage{booktabs}       % professional-quality tables
\usepackage{amsfonts}       % blackboard math symbols
\usepackage{nicefrac}       % compact symbols for 1/2, etc.
\usepackage{microtype}      % microtypography
\usepackage{xcolor}         % colors
\usepackage{graphicx}
\usepackage{amsmath}

\usepackage{multirow}

\usepackage[shortlabels]{enumitem}
\usepackage{tcolorbox}
\tcbuselibrary{breakable}
\usepackage{makecell}
\usepackage{wrapfig}
\usepackage{caption}
\usepackage{enumitem}
\usepackage{listings}
\usepackage{pdfpages}

\usepackage[preprint]{neurips_2025}

\usepackage[utf8]{inputenc} % allow utf-8 input
\usepackage[T1]{fontenc}    % use 8-bit T1 fonts
\usepackage{hyperref}       % hyperlinks
\usepackage{url}            % simple URL typesetting
\usepackage{booktabs}       % professional-quality tables
\usepackage{amsfonts}       % blackboard math symbols
\usepackage{nicefrac}       % compact symbols for 1/2, etc.
\usepackage{microtype}      % microtypography
\usepackage{xcolor}         % colors

\title{ActionSink: Toward Precise Robot Manipulation with Dynamic Integration of Action Flow}

\author{%
  Shanshan Guo$^1$$^*$, Xiwen Liang$^2$$^*$, Junfan Lin$^3$, Yuzheng Zhuang$^4$,\\ \textbf{ Liang Lin$^2$, Xiaodan Liang$^2$$^3$$^\dagger$}
  \\
  $^1$Northeastern University, $^2$Shenzhen Campus of Sun Yat-Sen University, \\
   $^3$Pengcheng Laboratory, $^4$Huawei Noah's Ark Lab
}

\begin{document}

\maketitle
\let\thefootnote\relax\footnotetext{$^*$Equal contribution, $^\dagger$Corresponding author}

\begin{abstract}
    Language-instructed robot manipulation has garnered significant interest due to the potential of learning from collected data. While the challenges in high-level perception and planning are continually addressed along the progress of general large pre-trained models, the low precision of low-level action estimation has emerged as the key limiting factor in manipulation performance. To this end, this paper introduces a novel robot manipulation framework, i.e., ActionSink, to pave the way toward precise action estimations in the field of learning-based robot manipulation. As the name suggests, ActionSink reformulates the actions of robots as action-caused optical flows from videos, called “action flow”, in a self-supervised manner, which are then used to be retrieved and integrated to enhance the action estimation. Specifically, ActionSink incorporates two primary modules. The first module is a coarse-to-fine action flow matcher, which continuously refines the accuracy of action flow via iterative retrieval and denoising process. The second module is a dynamic action flow integrator, which employs a working memory pool that dynamically and efficiently manages the historical action flows that should be used to integrate to enhance the current action estimation. In this module, a multi-layer fusion module is proposed to integrate direct estimation and action flows from both the current and the working memory, achieving highly accurate action estimation through a series of estimation-integration processes. Our ActionSink framework outperformed prior SOTA on the LIBERO benchmark by a 7.9\% success rate, and obtained nearly an 8\% accuracy gain on the challenging long-horizon visual task LIBERO-Long. %We also conducted extensive evaluations on Franka Kitchen and Meta-world, where our method consistently surpassed the baseline approaches.
% {\color{red}{The experimental results have proven that ...}}
\end{abstract}    
\section{Introduction}
\label{sec:intro}

\label{sec:intro}

\begin{figure*}[!t]
    \centering
    \includegraphics[width=\textwidth]{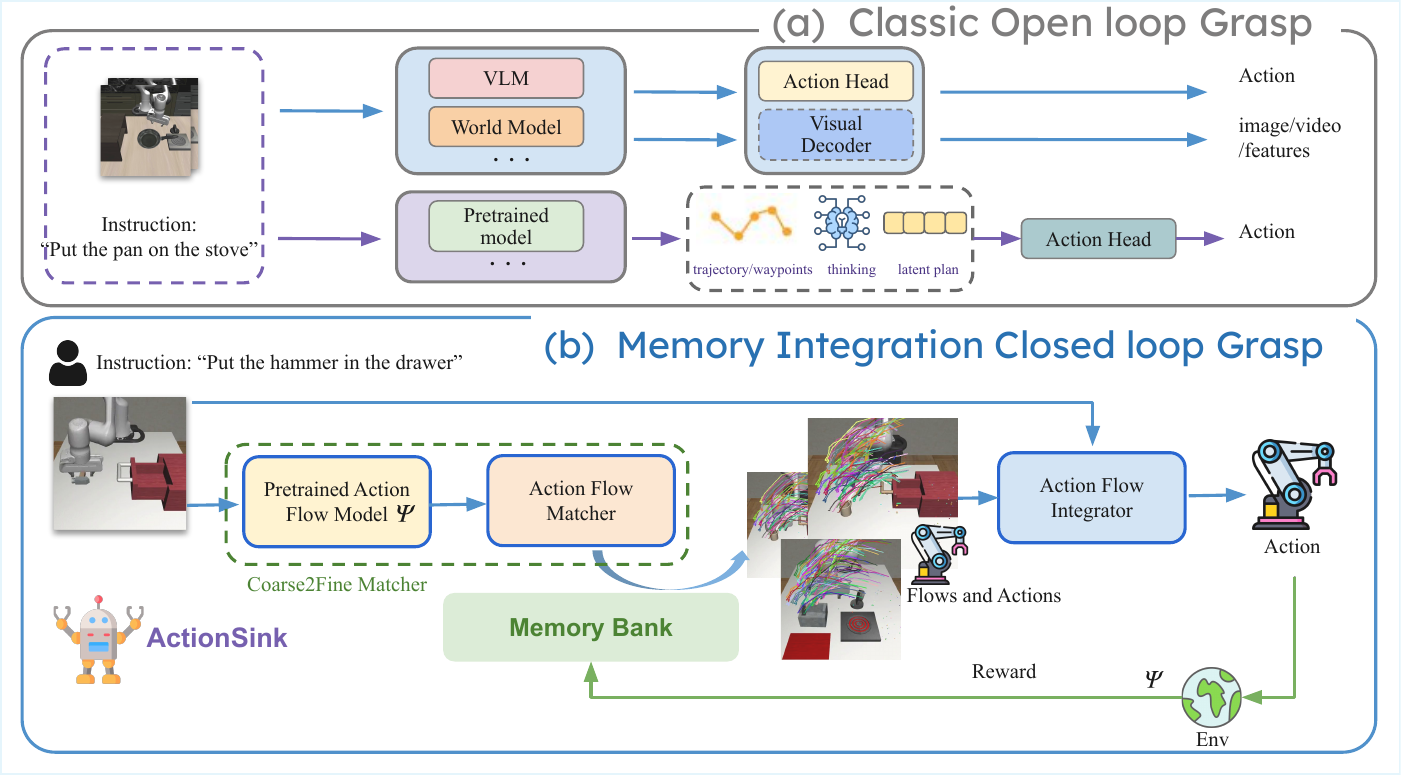}
    \caption{Comparison of ActionSink with other models. (a) Classic open loop grasp. This includes using VLM and World model as the backbone network to predict actions or specific features (\textit{e.g.}, RT-2 \cite{zitkovich2023rt2}, RT-X \cite{vuong2023openrt-x}, RT-H \cite{belkhale2024rt-h}, 
    %and
    VILA \cite{hu2023look}, Octo \cite{octo_2023}, Surfer \cite{Ren2023SurferPR}, Daydreamer \cite{wu2023daydreamer}, 
    %and
    3D-VLA \cite{zhen20243d}, {\it etc}.), or obtaining intermediate guidance including trajectory, waypoints, reasoning or potential plans through two-stage training (e.g., \cite{wang2023mimicplay,gu2023robotic,bharadhwaj2024track2act,wen2023any,Yuan2024GeneralFA,Xu2024FlowAT}, {\it etc}.). These methods all have problems with computational redundancy or the gap between high-dimensional guidance and low-level estimation. (b) ActionSink is a close loop grasping model with reward selection memory mechanism and dynamic integration of action flow. It can combine direct estimation of high-dimensional observations and correction of low-dimensional actions.}
    \vspace{-1.3em}
    \label{fig:comparision}
\end{figure*}

Recent advances in interactive artificial intelligence have heightened interest in robot manipulation tasks~\cite{bharadhwaj2024track2act,belkhale2024rt,dalal2024plan,escontrela2024video,du2024learning,ajay2024compositional,wang2023mimicplay,kumar2024robohive,opensora}. Many researchers are using deep learning to handle complex instructions and image understanding, and they usually use the collected robot trajectory data for end-to-end learning from instructions, observations to actions~\cite{belkhale2024rt,bahl2023affordances,nair2022r3m,zhang2024affordance}.

Robot manipulation constitutes a process mapping high-dimensional physical observations to low-dimensional actions. The substantial gap between these representations makes accurate action estimation challenging, necessitating an effective bridge connecting high-dimensional perception to low-dimensional control.

While visual language models (VLMs) have shown remarkable success, most methods use them as backbone networks (Fig.~\ref{fig:comparision}a), often adding visual decoders to predict images or videos as action constraints. For general manipulation model, this introduces action-irrelevant computational redundancy, fundamentally due to the absence of an optimized connection between high-dimensional observations and low-dimensional actions.

Conventionally, robotic manipulation can be roughly divided into high-level planning and low-level control. Current research focuses on enhancing high-level perception (Fig.~\ref{fig:comparision}a) by predicting trajectories, waypoints, thinkings or latent plans to bridge this gap~\cite{bahl2023affordances,wang2023mimicplay,wen2023any}. Nevertheless, low-level control remains the critical performance bottleneck~\cite{wang2023mimicplay,black2023zero}. Recent approaches like Im2Flow2Act~\cite{Xu2024FlowAT} introduce object-centric flow representations, while General-flow~\cite{Yuan2024GeneralFA} employs language-conditioned flow prediction for scalable learning.

% Unlike prior work, we aim to explore the fundamental disconnect between high-dimensional observations and precise low-dimensional action estimation. 
Unlike prior work, we aim to explore a method to better connect high-dimensional observation data with accurate estimation of low-dimensional actions.
Direct end-to-end estimation inevitably introduces deployment errors that cannot be resolved by additional training data alone. Inspired by traditional machine contro, the PID~\cite{ziegler1942optimum}
method introduces an integral term to dynamically eliminate the PD error. And recent retrieval-augmented generation techniques~\cite{zhu2024retrieval}, We propose bridging this gap through a transformation pipeline: \textit{high-dimensional observation} $\rightarrow$ \textit{action flow} $\rightarrow$ \textit{integrated matching of historical actions with direct estimation} $\rightarrow$ \textit{low-dimensional action}.

Specifically, we introduce action flow—a novel optical flow representation extracted from unlabeled videos in a self-supervised manner. Observational data often contains task-irrelevant information, such as background elements or object motions not caused by robot actions (e.g., human feet in the background). Directly fitting these factors forces models to memorize spurious correlations, making it difficult to capture the true environment dynamics relevant to task execution. Action flow exclusively encodes pixel changes caused by robot actions relative to language commands, creating a compact, noise-resistant action space. However, matching methods often suffer from noise sensitivity and limited diversity in complex observation spaces.

To address the aforementioned challenges, we propose \textit{ActionSink}, a novel memory-integrated closed-loop robot manipulation model illustrated in Fig.~\ref{fig:comparision}(b). As its name suggests, ActionSink integrates action-specific optical flows to estimate target actions, significantly improving data utilization efficiency while maintaining action estimation diversity and accuracy. The framework comprises two core components: a \textbf{Coarse-to-Fine Action Flow Matcher} that refines the current action flow estimate through iterative matching with similar historical flows and their corresponding actions, overcoming diversity limitations while enhancing matching precision; and a \textbf{Dynamic Action Flow Integrator} that combines matching results with current direct feature-space estimations to generate precise actions while maintaining a working memory pool. This memory pool dynamically updates by adding new action flows/actions and pruning obsolete memories, ensuring real-time efficiency.

ActionSink achieves a 68\% average success rate on LIBERO (+7.9\% relative improvement) and nearly 8\% accuracy gain on the challenging LIBERO-Long benchmark.

\noindent Our contributions can be summarized as follows:
\begin{itemize}
    \item We propose ActionSink, the first robot manipulation model that dynamically integrates action flows through coarse-to-fine matching and memory-based integration, bridging the gap between high-level planning and low-level actions to achieve precise control.
    
    \item We develop an action flow representation paradigm decoupled from optical flow, constructing a language-driven compact action space.
    
    \item We conduct comprehensive evaluations on LIBERO~\cite{liu2024libero}, Franka Kitchen~\cite{fu2020d4rl}, and Metaworld~\cite{yu2020meta}, demonstrating ActionSink's superior performance over strong baselines across diverse settings.
\end{itemize}
\section{Related Work}

\textbf{Classical Approaches to Long Horizon Robotics:} Significant progress has been made in leveraging foundational models for long-horizon robotic tasks. For instance, Hip~\cite{ajay2024compositional} combines foundational models into a hierarchical system, while SayCan~\cite{ahn2022can} uses LLMs to decompose tasks into skill-based value functions for low-level control. Plan-Seq-Learn~\cite{dalal2024plan} employs LLM-guided strategies to direct RL-based policies without predefined skills. Unipi~\cite{du2024learning} utilizes large-scale internet data for text-conditioned video generation, and MimicPlay~\cite{wang2023mimicplay} leverages human gameplay data for latent plan learning to guide motor control. In contrast, our approach focuses on recalling past experiences from memory to inform the current learning process.
% \\

\textbf{Policy Conditioning Representations:} 
In robot policy learning, language-conditioned policies \cite{brohan2022rt,ahn2022can} use free-form language as task specifications, but these are often insufficiently descriptive, leading to ambiguity in mapping instructions to actions. Another approach leverages large-scale text-to-image models \cite{gao2024can, du2024learning}, conditioning policies on goal images for reinforcement learning. However, pre-trained diffusion models tend to predict pixel-level information for entire images, which is computationally complex, redundant, and lacks dynamic context, making them less suitable for tasks like grasping. Multimodal methods embed both task-conditioned text and goal-conditioned images together \cite{xiao2022robotic,jiang2022vima}, aligning actions with textual and visual cues. Affordance-based policies \cite{bahl2023affordances} use visual affordance models to estimate potential human interactions within a scene, enabling robots to recognize the functional possibilities of objects. Recently, trajectory-guided conditional policies have emerged as a promising approach, where providing or inferring trajectory sketches makes task generalization feasible \cite{wang2023mimicplay,gu2023robotic}. Some of these methods \cite{bharadhwaj2024track2act,wen2023any} pre-train models to predict the future motion of arbitrary points in the observation or predict latent high-level plans to guide low-level actions. In contrast, we pre-train sparse local optical flow specifically for the robot arm and target objects, offering more precise movement predictions and better capability to capture dynamic motion features.
% \\

\textbf{Retrieval-Augmented Generation (RAG):}
Retrieval-augmented generation (RAG) was first introduced in the field of natural language processing (NLP)\cite{miao2024integrating,zhu2024retrieval,chen2024benchmarking,he2024large}, where it combines retrieval and generation modules to enhance a model's ability to handle previously unseen inputs. In the context of robotic grasping, the potential of RAG remains largely underexplored. The RAG model can retrieve relevant grasp strategies or environmental information for the current task and dynamically generate adaptive grasp strategies using a generative model. This combination of retrieval and generation gives RAG greater generalization capability and robustness when dealing with complex grasping tasks, especially in scenarios involving unseen objects or dynamic environments.
In recent years, several works have begun to explore the enhancement of grasp strategy generation through the retrieval of relevant historical data. However, most of these methods\cite{zhu2024retrieval} rely heavily on the direct application of retrieval results, lacking the ability to dynamically generate strategies for new scenes. Our method integrates direct estimation with Retrieval-Augmented Generation to adjust the estimation process, and we propose the `action flow' that enables dynamic generation capabilities tailored to the current scene.

\label{sec:related}

\section{ActionSink}
\label{sec:method}
\begin{figure*}
    \centering
    \includegraphics[width=\textwidth]{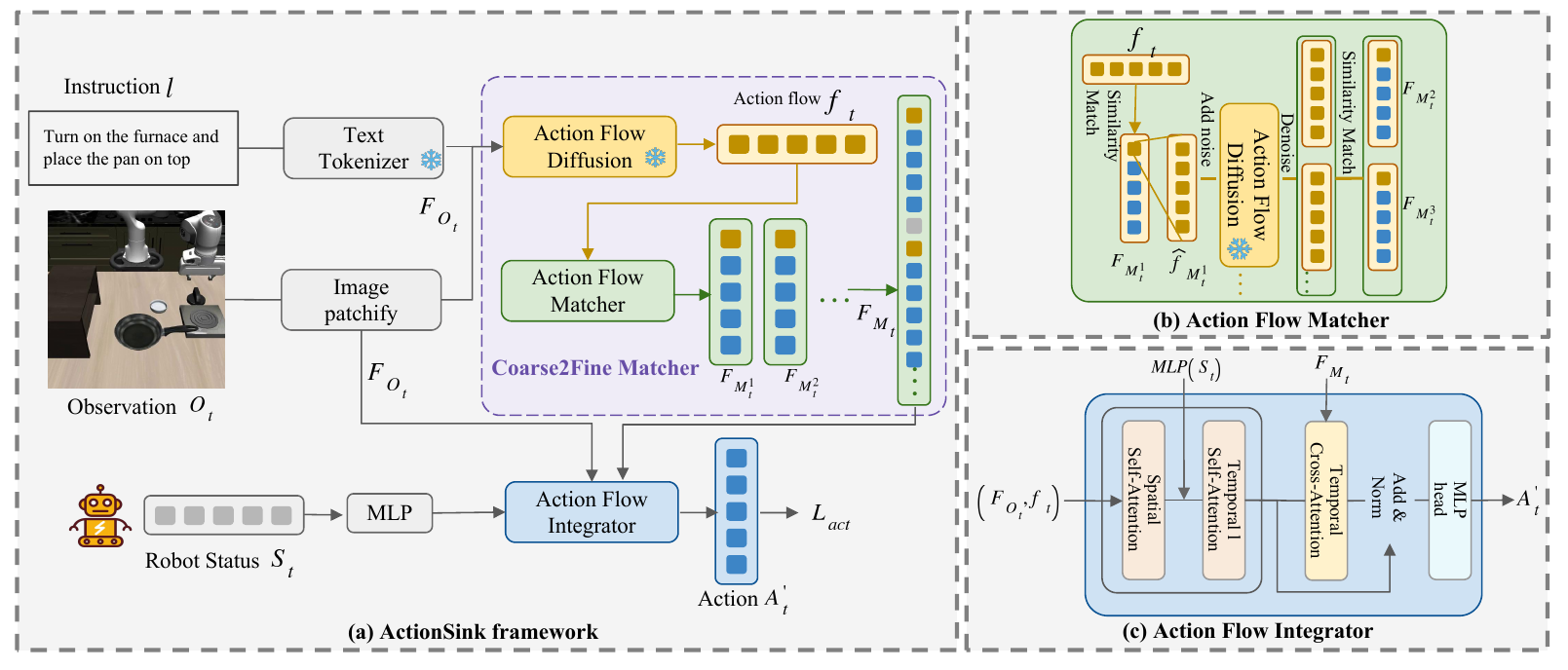}
    \caption{ActionSink Overview. It mainly consists of Coarse2Fine Matcher and an Action Flow Integrator. In Coarse2Fine Matcher, we first use the pre-trained Action Flow Diffusion to generate the current Action flow for coarse-to-fine matching through Action Flow Matcher. Then the Action Flow Integrator fuses the direct estimation of the current observation and the correction of the matching, and finally predicts and executes the action.}
    \label{fig:framework}
    \vspace{-2mm}
\end{figure*}
ActionSink aims to bridge high-dimensional observations to precise low-dimensional actions through a transformation pipeline: high-dim input (Sec.~\ref{sec:3.1}) \(\rightarrow\) action flow extraction and matching (Sec.~\ref{sec:3.2.1}) \(\rightarrow\) integrated historical-direct estimatio (Sec.~\ref{sec:3.2.2}) \(\rightarrow\) low-dim action.
As shown in Figure\ref{fig:framework}, we introduce a robot manipulation model with Coarse-to-Fine Action Flow Matcher that gets the current action flow and various historical matching action flows, then Dynamic Action Flow Integrator combines current estimates with historical corrections. And a working memory pool with reward-based retention (Sec.~\ref{sec:3.3}) dynamically maintains relevant action flows.
ActionSink aim to bridge high-dimensional observations to precise low-dimensional actions.

\subsection{Inputs and Outputs}
\label{sec:3.1}
We provide a detailed description of the inputs and outputs of ActionSink in Figure\ref{fig:framework}(a) as follows:
\begin{itemize}[leftmargin=1em]
\item\textbf{Memory Bank.} We denote the memory bank as $L_i = \{(E_i, V_i, F_i, Act_i)\}_{i=1}^{N}$, where $N$ denotes the total number of episodes, the instruction embedding $E_i$, the first frame image $V_i$, the action flow $F_i$, and the action label from the previous episode $i$ $Act_i$.
\item\textbf{Language input.} For 
 the language instructions $l$, we use a pre-trained CLIP-Text encoder \cite{radford2021learning} to obtain embeddings $E_{text}$.

\item\textbf{Visual input.} 
For visual observation of RGB image $O_t$, which is divided into patches $F_{O_t} \in \mathbb{R}^{patches \times C}$
% which are combined with $Z$ as conditions for the diffusion model. 

\item\textbf{Robot state input.} The robot state includes 6 dimensions of robot arm movement S =
($x$, $y$, $z$, roll, pitch, yaw). We use linear layers to encode them.

\item\textbf{Outputs.} The output of ActionSink is the intermediate output $f_t \in \mathbb{R}^{K \times N \times 2}$ in the Coarse2Fine process, where $N$ is the number of action flow pixels and $K$ is the length of action flow. And the robot action $A'_t$ predicted by the final output action stream integrator. The action $A'_t$ contains the incremental state S of the robot end effector and the binary state $G \in \mathbb(0, 1)$ of the gripper, that is $A'_t = (S, G) \in \mathbb{R}^{1 \times 7}$.
\end{itemize}

\subsection{ActionSink Architecture}
\subsubsection{Coarse-to-Fine Action Flow Matcher}
\label{sec:3.2.1}
The currently observed action flow is obtained through the pre-trained action flow diffusion model, and a preliminary match is obtained through coarse retrieval. Then, more diverse and similar action flows are generated and matched through Denoising for fine retrieval.
\begin{itemize}[leftmargin=1em]
\item\textbf{Action Flow Diffusion.}
We model robotic action dynamics as a local optical flow prediction task. A conditional diffusion model $\mathcal{\psi}$ is pre-trained on unlabeled video data to predict $K$-step action flow within task-relevant regions (e.g., robot arm-object interactions), conditioned on RGB observations $O_t$ and language instructions $l$. As shown in Figure~\ref{fig:framework}(a), $\mathcal{\psi}$ generates task-specific action flow $f_t$ by:
\begin{equation}
f_t = \mathcal{\psi}(F_{O_t}, E_{text}; \theta)
\end{equation}
where $\theta$ denotes learned parameters that couple visual perception with language.
\item\textbf{Coarse retrieval.} 
The retrieval process computes multi-modal similarity scores between the current input and stored memory segments. For each segment \( m \), the aggregated score \( S_m \) integrates three components with learnable weights:
\begin{equation}
S_m = 
\underbrace{\lambda_1 S_I}_{\mathclap{\scalebox{0.63}{\text{(Instruction)}}}} + 
\underbrace{\lambda_2 S_V}_{\mathclap{\scalebox{0.63}{\text{(Observation)}}}} + 
\underbrace{\lambda_3 S_F}_{\mathclap{\scalebox{0.63}{\text{(Action Flow)}}}}
\end{equation}
where instruction similarity \( S_I \) and observation similarity \( S_V \) are computed via cosine similarity between embeddings, while action flow similarity \( S_V \) is measured through sliding window alignment and latent space distance. The observation term \( S_V \) ensures environmental style consistency, whereas the action flow term \( S_F \) serves as the primary matching target. The top-ranked memory segment provides both the matched action flow \( \hat{f}_m \) and associated actions \( \hat{A}_m \), yielding the coarse retrieval result $F_{M_t^1}=[\hat{f}_{M_t^1}, \hat{A}_{M_t^1}]$

\item\textbf{Denoising for fine retrieval.} To enhance the diversity and accuracy of matching action flows, we employ a denoising diffusion process with iterative retrieval. As shown in Figure~\ref{fig:framework}(b), we first extract the action flow \(\hat{f}_{M_t^1}\) from coarse match\(F_{M_t^1} = [\hat{f}_{M_t^1}, \hat{A}_{M_t^1}]\), then inject noise and perform backward diffusion through model \(\psi\) to generate novel samples \(\{f_t'\}\) that preserve semantic relevance while deviating from \(\hat{f}_{M_t^1}\). The generation follows:

\begin{equation}
f_t' = \sqrt{\alpha_t} \hat{f}_{M_t^1} + \sqrt{1 - \alpha_t} \epsilon, \quad \epsilon \sim \mathcal{N}(0,1), \ t \in T_{\text{steps}}, \ \alpha_t \in [0,1]
\end{equation}

where \(T_{\text{steps}}\) denotes denoising iterations and \(\alpha_t\) controls noise scheduling. After generating \(n-1\) diversified candidates via this process, we iteratively retrieve their corresponding matches. By interleaving state markers and concatenating all refined results, we construct the final matching result \(F_{M_t}=(F_{M_t^1},F_{M_t^2},\dots,F_{M_t^n})\)

\end{itemize}

\subsubsection{Dynamic Action Flow Integrator}
\label{sec:3.2.2}
In this module, we aim to combine the matching results with the direct estimation of the current action flow feature space to help perform more accurate actions in the current task.

\textbf{Fusion of direct estimation and action integration. } As shown in Figure\ref{fig:framework}(a), we take the robot's observation image $O_t$, action flow $f_t$, robot state $S_t$, and matching result $F_{M_t}$ for integration as input to predict the action $A_t^\prime$ that the robot should take at time $t$. Therefore, the prediction process of action $A_t^\prime$ can be expressed as:
\begin{equation}
    A'_t = \Phi(F_{O_t},f_t,F_{M_t}, \text{MLP}(S_{t})).
\end{equation}
The model details of the action flow integrator are shown in Figure~\ref{fig:framework}(c). The direct estimation consists of a spatial self-attention layer and a temporal self-attention layer. In order to effectively integrate the matching results $F_{M_t}$ into the network, we utilize the cross-attention mechanism. This approach enables the model to extract similar and different action flow and action information from the matching results. We fuse and normalize the output of the direct estimation with the output of the action integration to ensure that the action is consistent with the current spatial and proprioceptive state without losing any information. Finally, it passes through a fully connected network (MLP).

\subsection{Reward for Selective Preservation}
\label{sec:3.3}
To optimize retrieval efficiency, we enforce memory diversity by discarding redundant actions with high relevance score similarity. Leveraging a pre-trained action flow model \(\psi\) as a reward generator \cite{escontrela2024video}, we design a dual-reward mechanism:

\vspace{-1\baselineskip}  % 缩小上方间距
\begin{equation}
r_{\text{flow}}^t = \ln p_{\theta} \left( f_{t} \mid O_t,l \right)
\label{eq:flow_reward}
\end{equation}

\vspace{-1\baselineskip}  % 缩小公式间间距
\begin{equation}
R_{\text{flow}} = \frac{1}{T-1} \sum_{t=1}^{T-1} r_{\text{flow}}^t
\label{eq:flow_quality}
\end{equation}

\vspace{-0.5\baselineskip}  % 继续缩小间距
\begin{equation}
R_{\text{total}} = \alpha R_{\text{flow}} + \beta R_{\text{action}}
\label{eq:total_reward}
\end{equation}

\vspace{-0.5\baselineskip}
Here \(R_{\text{action}}\) evaluates action success rates, while \(R_{\text{flow}}\) reflects visual performance via action flow prediction likelihood. Episodes are stored only if \(R_{\text{total}} > \tau_{\text{total}}\), ensuring the memory pool retains visually consistent and grasps successful actions.

\subsection{Loss}
For the action flow diffusion in coarse2fine matcher module, during the forward diffusion process, noise $\epsilon \sim \mathcal{N}(0, 1)$ is added to the action flow $f_t$ at step $t' \in [1, k]$. The denoising function $\epsilon_\theta$ is trained to minimize the loss:

\begin{equation}
    \mathcal{L}_{\text{MSE}} = \left\lVert \epsilon - \epsilon_\theta \left( \sqrt{1 - \beta_{t'}} \cdot f_t + \sqrt{\beta_{t'}} \cdot \epsilon \mid {t'}, [F_{O_t}, E_{text}] \right) \right\rVert^2,
\end{equation}
Where $\beta_t$ is the noise schedule at step $t$. For action prediction loss $L_{act}$,  we use Cross Entropy Loss to calculate the loss between the predicted action $A_t^\prime$ and the ground truth action $A_t$.

\begin{table}[t]
    \centering
    % \vspace{-4.5em}
    \caption{Performance comparison on different Libero benchmarks.}
    \vspace{0.35em}
    % \scriptsize
    \resizebox{0.95\linewidth}{!}{
    \begin{tabular}{l|c|cccc|c}
    \toprule
        Model & Training Data & Libero-Spatial &  Libero-Object &  Libero-Goal & Libero-Long & Mean \\ \midrule
        BC~\cite{brown2020language} & Lang &39.00 & 51.83 & 42.50 & 16.67 & 37.50 \\
        R3M-BC~\cite{nair2022r3m} & All & 49.17 & 52.83&  5.33 &  9.17 & 29.13 \\
        Unipi~\cite{du2024learning} & All & 69.17 & 59.83 & 11.83 & 5.83 & 36.67\\
        ATM~\cite{wen2023any} & All & 68.50 & 68.00 & 77.83  &  39.33 & 63.42\\ 
        % Im2Flow2Act & & & \\
        % General-Flow & & &\\ 
        \midrule
        ActionSink & All & 72.83 & 72.67 & 80.00 & 47.00 & 68.13\\
        
    \bottomrule
    \end{tabular}
    }
    \vspace{-2mm}
    \label{tab:libero}
\end{table}
\section{Experiment}
\label{sec:exp}
% describe
We conduct experiments on challenging simulators and benchmarks such as LIBERO, Franka Kitchen, and Metaworld.
Our experiments aim to address four key questions: 1) How accurate is ActionSink in executing various simple and complex language instructions? 2) Which modules of ActionSink play a key role in its performance? 3) Is the robot capable of performing long-view tasks with real-time working memory updates? 4) How robust and generalizes is ActionSink in non-distributed scenarios?

\subsection{Experiment Settings}
In this section, we describe the experimental setup used to evaluate our proposed method. All training and testing uses a fixed third-person view to capture the entire scene. Training is performed on a workstation equipped with eight NVIDIA RTX 4090 graphics card to accelerate the training process of the neural network. For the action flow integrator, we use 20\% of the dataset for training. For each task, we record the task success rate, which is the percentage of the robot successfully completing the task. To ensure fairness, all methods are tested under the same experimental conditions. In LIBERO, we evaluate our method by running 20 rollouts for each task. In Franka Kitchen and Metaworld, we run 40 rollouts for each task to evaluate. No additional noise is introduced in the experiment, and the robot state is assumed to be fully observable.

\subsection{Benchmark and Baselines}
\textbf{Simulation Benchmarks.} We conducted experiments on three widely used benchmarks (LIBERO, Franka Kitchen, and Metaworld), each of which provides different challenges for robotic manipulation tasks. We evaluated our method on both single-task and multi-stage long-term manipulation tasks. The LIBERO dataset encompasses diverse robotic challenges, such as object manipulation and long-term planning, with specific benchmarks including Libero-Spatial, Libero-Object, Libero-Goal, and Libero-Long. Each test set contains 10 tasks, and each instruction in these benchmarks usually contains 2-4 subtasks. Franka Kitchen and Metaworld focus on object interaction and single-step robotic manipulation in kitchen environments and daily life.\\
\textbf{Action Flow Data Collection.} For the action flow diffusion in the Coarse2Fine matcher module, the training data is generated using the ground truth optical fineflow of Cotracker3 \cite{karaev2024cotracker3}. To focus on the pixel changes caused by the robot's movements, we use GroundingDINO \cite{ren2024grounding} and SAM2 \cite{ravi2024sam} to extract the dynamic optical flow of the robot and related objects. This ensures that irrelevant background noise is excluded, thereby improving the integration effect with the action flow integrator.

\textbf{Baseline.} We selected BC, R3M-BC, Unipi, Mimicplay, ATM and RAEA as the baseline models for ActionSink.
\begin{itemize}[leftmargin=1em]
    \item \textbf{Language conditioned policy.} Vanilla behavior cloning (\textit{BC}~\cite{brown2020language}) baseline trained on a set of action-labeled expert demonstrations where language serve as task specifications. Additionally, (\textit{R3M-BC}~\cite{nair2022r3m}) is a goal-conditioned \textit{BC} variant that utilize \textit{R3M} pre-trained visual representations.
    \item \textbf{Video conditioned policy.} \textit{Unipi}~\cite{du2024learning} is a text-conditioned video diffusion model that generates a video plan from an initial frame and language instructions. 
    \item \textbf{Trajectory
    conditioned policy.} \textit{ATM}~\cite{wen2023any} pre-trains the model to track any points in a video, while \textit{MimicPlay}~\cite{wang2023mimicplay} derives high-level latent plans from human demonstration videos, enabling long-term grasping tasks. 
    % Im2Flow2Act introduce an object-centric flow representation for action modeling.
    % We implemented \textit{RT-trajectory}, which uses a more compact, coarse trajectory sketch as policy conditioning. However, this coarse trajectory requires additional data input rather than being predicted by the model itself. 
    \item \textbf{Memory-based policy.} We compared \textit{RAEA}~\cite{zhu2024retrieval}, which retrieves relevant policies from an external repository through a straightforward storage and retrieval mechanism.
\end{itemize}

\begin{table}[t]
    \centering
    % \vspace{-4.5em}
    \caption{Comparison of performance in Franka Kitchen environments.}
    % \vspace{-0.5em}
    % \scriptsize
    \resizebox{0.85\linewidth}{!}{
    \begin{tabular}{l|ccccc|c}
    \toprule
        \multirow{2}{*}{Method} & \multicolumn{5}{c|}{Franka Kitchen} & \multirow{2}{*}{Mean}\\
        \cmidrule(lr){2-6} 
        & K-Slide & K-hinge & K-Burner & K-Light & K-Microwave\\ \midrule
        R3M-BC~\cite{nair2022r3m} & 52.5 & 87.5 & 42.5 & 77.5 & 30.0 & 58.0 \\
        ATM~\cite{wen2023any} & 62.5 & 85.0 & 65.0 & 85.0 & 70.0 & 73.5\\
        MimicPlay~\cite{wang2023mimicplay} & 55.0 & 80.0 & 50.0 & 80.0 & 50.0 & 63.0\\ 
        RAEA~\cite{zhu2024retrieval} & 50.0 & \textbf{100.0} & 65.0 & 92.5 & 45.0 & 70.5\\ 
        % \rowcolor{gray!15}
        % Im2Flow2Act & & & & &\\
        ActionSink & \textbf{80.0} & 92.5 & \textbf{92.5} & \textbf{95.0} & \textbf{90.0} & \textbf{90.0}\\ 
        
    \bottomrule
    \end{tabular}
    }
    \vspace{-2mm}
    \label{tab:frankakitchen}
\end{table}

\begin{table}[t]
    \centering
    % \vspace{-4.5em}
    \caption{Comparison of performance in Meta-World environments.}
    % \vspace{-0.5em}
    % \scriptsize
    \resizebox{0.95\linewidth}{!}{
    \begin{tabular}{l|ccccc|c}
    \toprule
        \multirow{2}{*}{Method} & \multicolumn{5}{c|}{Meta-World}& \multirow{2}{*}{Mean}\\
        \cmidrule(lr){2-6} 
        & Assembly & Bin-picking & Botton-press & Drawer-close & Hammer \\ \midrule
        R3M-BC~\cite{nair2022r3m} & 37.5 & 22.5 & 22.5 & 60.0 & 72.5 & 43.0\\
        ATM~\cite{wen2023any} & 57.5 & 65.0 & 70.0 & 75.0 & 72.5 & 68.0\\
        MimicPlay~\cite{wang2023mimicplay} & 47.5 & 37.5 & 57.5 & 75.0 & 62.5 & 56.0\\ 
        RAEA~\cite{zhu2024retrieval} & 57.5 & 50.0 & 42.5 & 85.0 & 87.5 & 64.5\\ 
        % Im2Flow2Act & & & & &\\ \midrule
        % \rowcolor{gray!15}
        ActionSink & \textbf{77.5} & \textbf{80.0} & \textbf{75.0} & \textbf{87.5} & \textbf{90.0} & \textbf{82.0}\\ 
    \bottomrule
    \end{tabular}
    }
    \vspace{-2mm}
    \label{tab:metaworld}
\end{table}

% \textbf{Evaluation: }
\subsection{Results on Robotic Manipulation}

% \\

\textbf{Experimental results on Franka Kitchen and Meta-World (Single Task): }
We evaluate 10 task categories on Franka Kitchen and Meta-World, with only 10 demonstrations per task. The results are shown in Tables \ref{tab:frankakitchen} and Table \ref{tab:metaworld}, 
and ActionSink achieves 90\% and 82.5\% success rates respectively, significantly outperforming the best baselines (16.5\% and 14\% improvement).
This demonstrates the effectiveness of dynamic integration in improving prediction accuracy. In the Franka Kitchen environment, RAEA performs strongly in the K-hinge and K-light tasks, but its performance is not stable, which may be related to directly retrieving and using information, as it is highly sensitive to retrieval accuracy. On the other hand, the ATM method shows relatively stable performance but has difficulty reaching peak efficiency. This suggests that pixel-level motion can better characterize robot grasping and low-level control, but it may be hindered by background noise and irrelevant information.

\textbf{Experimental results on LIBERO (Long Horizon):} 
We present the main results of LIBERO in Table ~\ref{tab:libero} . It is evident that by combining the Coarse2Fine Matcher and the Action Flow aggregator, our method ActionSink achieves an average success rate of 68\%, a relative improvement of 7.9\% over the previous best average success rate of 63\%. 
Comparison with ATM shows that the motion caused by the robot's actions and the visual changes directly driven by these actions reflect the physical reality of dynamic interactions. We call this efficient representation Action Flow. 
The failure of the video prediction model Unipi is mainly because the videos are more complex and redundant, lacking direct physical features related to the actions, which makes them less applicable for low-level control tasks.
Notably, we achieve an improvement of nearly 8\% on the most challenging Libero Long task with only a small amount of labeled data, which demonstrates that our method offers great potential for complex and long-term tasks due to its ability to update the memory pool.

\subsection{Generalization Ability}
As shown in Table\ref{tab:gener}.We also conducted generalization experiments. Specifically, we tested Libero-Spatial, Libero-Object, and Libero-Goal with the model trained on Libero-Long, denoted as Unseen1, Unseen2, and Unseen3. The migration scenes include different lighting and background environments, unseen objects (e.g., wine bottles, cream cheese, ketchup), and more distractions (e.g., more objects). ActionSink still maintains a higher success rate than other models, which shows that the Action Flow integrator can effectively fuse similar historical matches to improve accuracy and maintain good generalization ability in scene changes.

\begin{table}[t]
  \renewcommand{\arraystretch}{1.2}
  \begin{minipage}[t]{0.48\textwidth} 
    \centering
    \caption{Ablation studies of different components and models in metaworld}
    \vspace{0.2em}
    \setlength{\tabcolsep}{1.8pt}
    \renewcommand{\arraystretch}{1.2}
    {   
        \begin{tabular}{l|cc}
            \toprule
            Model            & Action Flow & ATM \\
            \midrule
            STA block        & 72.5   & 68.0  \\
            Matcher+Integrate & 82.0     & 76.5 \\
            \bottomrule
        \end{tabular}
        \label{tab:4_actionflow}
    }
  \end{minipage}
  \hfill 
    \begin{minipage}[t]{0.5\textwidth} 
    \centering
    \caption{Transfer experiments of scenes, tasks or objects outside of distribution.}
    \vspace{0.35em}
    % \scriptsize
    \resizebox{1\linewidth}{!}{
    \begin{tabular}{l|ccc|c}
    \toprule
        Model & Unseen 1 &  Unseen 2 &  Unseen 3 & Mean \\ \midrule
        R3M-BC~\cite{nair2022r3m} & 22.17 & 19.17 & 7.33 & 16.22  \\
        ATM~\cite{wen2023any} & 42.50 & 48.50 & 53.20 & 48.07 \\ 
        \midrule
        ActionSink & 59.13 & 55.60 & 67.50 & 60.74 \\
        \bottomrule
    \end{tabular}
    }
    \label{tab:gener}
    \vspace{-1em}
    \end{minipage}
\end{table}

\subsection{Ablation Study}
In this section, we will explore the key points in model design. Specifically, we will discuss the effects of action flow, Coarse2Fine iterative matching, Action Flow Integrator,  selective reward retention, memory and inference time, and matching parameters on the performance of ActionSink. We designed a series of ablation experiments and made some assumptions and experiments: \textit{(i)} We use ground-truth (GT) action flow to directly match and integrate the real action flow, we use different modal data for matching. \textit{(ii)} Direct matching instead of Coarse2Fine iterative estimation and matching. \textit{(iii)} Remove the action flow integrator, that is, directly estimate from the current action flow. \textit{(iv)} Randomly retain rewards according to task type \textit{(v)} Adjust the weight of the matching parameters. The results of these experiments are analyzed in detail below.
% \\

\textbf{Action flow.}
As shown in Table~\ref{tab:6_ablation_module}, the performance of matching and integrating with GT Action flow can reach up to 78.33\%, indicating that Action Flow can provide accurate action representation. We also evaluated the effect of matching using a combination of different modal data, and the results are shown in Table~\ref{tab:7_ablation_retrieve}. 
% Image-based matching has the worst effect, which may be because only one image reflecting the environment is saved, resulting in biased action understanding and being easily affected by irrelevant background information.
\\
\textbf{Coarse2Fine Matching.}
As shown in Table~\ref{tab:7_ablation_retrieve}, Coarse2Fine iterative estimation and matching outperforms direct matching, which is attributed to the sensitivity of Action Flow Integrator to diversity and the accuracy of matching.
\\
\textbf{Action Flow Integrator.}
Table~\ref{tab:6_ablation_module} shows that the action estimation accuracy drops sharply to 65.5\% after removing this component. Combined with the results in Table~\ref{tab:4_actionflow}, the performance of the Matcher+Integrate component in both ActionSink and ATM is significantly better than the STA module (temporal and spatial self-attention mechanism, i.e. direct estimation), and the action flow component outperforms the ATM in both configurations. This highlights the key role of this module combination in achieving optimal performance, and its action flow representation can provide a more robust estimation space.
\\
\textbf{Selective reward retention.} 
Retrieval and fine estimation require high efficiency and high accuracy of the working memory pool. For random retention based on the task type, regardless of its accuracy, query efficiency, or matching quality. As shown in Table~\ref{tab:6_ablation_module}, the SR dropped from 68.17\% to 65.50\%.
\begin{wrapfigure}{r}{0.45\textwidth}
    \centering
    \vspace{-0.8em}
    \caption{Parameter Study of \(\lambda\) in matching}
    \begin{tabular}{cccc}
        \hline
        \(\lambda_1\) & \(\lambda_2\) & \(\lambda_3\) & Success (\%) \\
        \hline
        0 & 0.143 & 0.887 & 71.17 \\
        0.33 & 0 & 0.667 & 72.33 \\
        0.75 & 0.25 & 0 & 64.5 \\
        0.3 & 0.1 & 0.6 & 72.83 \\
        \hline
    \end{tabular}
    \label{tab:para}
    \vspace{-1em}
\end{wrapfigure}
\textbf{Memory and inference time analysis.}
Due to the high reuse rate of robot actions, good performance can be maintained with a small memory. The experiments in Table \ref{tab:memory} show that when the pool size exceeds a certain value, the performance tends to stabilize, so that it can be flexibly adjusted to balance efficiency and scalability. The CPU memory usage remains relatively constant, and its sensitivity to the number of rounds is low.
In terms of reasoning time, the reasoning time also tends to stabilize after the initial stage, indicating that the system has reached a stable state in terms of computational efficiency
\\
\textbf{Parameter combination of matching.}
Figure~\ref{tab:para} shows that Actionsink is more sensitive to action flow( \(\lambda_3\)).The first frame image and language matching have little impact on the fluctuation of the results.

\begin{table}[t]
  \renewcommand{\arraystretch}{1.2}
  \begin{minipage}[t]{0.48\textwidth} 
    \centering
    \caption{Ablation study on the effect of different modules of the model on action prediction accuracy in Libero-Spatial.}
    \vspace{0.5em}
    \resizebox{1\linewidth}{!}{
    \begin{tabular}{c|l|c}
        \toprule
        Method & Status & Suc. Rate \\
        \midrule
        \multirow{4}{*}{Ours} 
          & Full & 72.83 \\
          & w GT Action Flow & 78.33 \\
          & w/o Action Flow Integrator & 65.50 \\
          & w/o Reward Retention & 68.17 \\
        \bottomrule
    \end{tabular}
    }
    \label{tab:6_ablation_module}
  \end{minipage}
  \hfill 
  \begin{minipage}[t]{0.5\textwidth} 
    \centering
    \caption{Ablation study on the matching of different modalities and different action flow matching methods in Libero-Spatial.}
    \vspace{0.5em}
    \resizebox{1\linewidth}{!}{
    \begin{tabular}{c|l|c}
        \toprule
        Method & Retrieve & Suc. Rate \\
        \midrule
        \multirow{5}{*}{Ours} 
          & Text & 64.17 \\
          & Image & 55.83 \\
          & Text+Image & 64.50 \\
          & Text+Image+Flow & 70.17 \\
          & Text+Image+Coarse2Fine Flow & 72.83 \\
        \bottomrule
    \end{tabular}
    }
    \label{tab:7_ablation_retrieve}
  \end{minipage}
  \vspace{-1.5em}
\end{table}

\begin{table}[t]
\centering
\caption{Inference cost and speed changes in different episodes} 
\resizebox{0.80\linewidth}{!}{
\begin{tabular}{l|ccccccc}
\toprule
    Episode & 1 & 10 & 25 & 50 & 100 & 150 & 200\\ \midrule
    GPU memory (MB) &923	&3104	&3890	&3892	&3891	&3892	&3892           \\
    CPU memory (MB) &5297	&5274	&5311	&5317	&5323	&5324	&5323           \\
    Inference Time (ms) &12	&56	&63	&65	&66	&66	&67            \\ \bottomrule
\end{tabular}
}
\label{tab:memory}
\vspace{-1.6em}
\end{table}
% \section{Discussion}
% \label{sec:discussion}

\section{Conclusion}
\label{sec:discussion}
We propose ActionSink, a novel robotic manipulation framework that dynamically integrates action flows, which significantly improves the accuracy of action estimation by converting robot action space into "action flow" and leveraging a dual-module approach (a coarse-to-fine matcher and a dynamic integrator with multi-layer fusion modules). It achieves state-of-the-art results on the LIBERO, Franka Kitchen, and Meta-world benchmarks. In addition, Action Sink's strong generalization ability in unseen scenes, instructions and tasks demonstrates the effectiveness of dynamically integrating action flows for low-level manipulation. These results highlight the potential of ActionSink in improving the accuracy of low-level actions in robotic manipulation.\\
% \textbf{Limitations}

\bibliographystyle{unsrt}
\bibliography{main}


%%%%%%%%%%%%%%%%%%%%%%%%%%%%%%%%%%%%%%%%%%%%%%%%%%%%%%%%%%%%

\appendix

% \section{Technical Appendices and Supplementary Material}
% Technical appendices with additional results, figures, graphs and proofs may be submitted with the paper submission before the full submission deadline (see above), or as a separate PDF in the ZIP file below before the supplementary material deadline. There is no page limit for the technical appendices.

%%%%%%%%%%%%%%%%%%%%%%%%%%%%%%%%%%%%%%%%%%%%%%%%%%%%%%%%%%%%

\newpage

\clearpage
\setcounter{page}{1}
% \maketitlesupplementary

\appendix

\section{Overview}
This supplementary document offers further information, results, and visualizations to complement the primary paper. Specifically, we encompass:
\begin{itemize}[left=0pt, label={•}] % Custom bullet point and no indentation
    \item Impacts and limitations statement;
    \item Simulation and tasks;
    \item Impact of segmentation effect;
    \item Implementation details;
    % \item Illustrative examples;
    % \item Complementary qualitative comparisons;
    \item More Experimental Results
    \item Visualizations of action flow prediction and action prediction.
\end{itemize}

\section{Impacts and limitations statement}
\label{Limitations}
\textbf{Potential Negative Social Impact.} 
Our approach does not have ethical risks in terms of dataset usage and privacy violation, as all benchmarks are public and transparent.\\
\textbf{Limitation and Future Work.}
1) Our model is based on 2D pixel-level action flow caused by robotic arm movements. While 2D data is cheaper and more readily available, it lacks spatial awareness, which is crucial since robotic manipulation occurs in 3D space. Future work could explore 3D action flow, providing the model with spatial transformation representations. This could potentially reduce the discrepancy between the input and output dimensional spaces and improve the accuracy of real-world manipulation.
2) The dynamic action flow integrator faces challenges when applied to long-horizon tasks. To address this, future research could focus on subgoal extraction, progress tracking, and matching mechanisms, which could enable faster and more responsive behaviors.
3) To enhance the model's action flow prediction capabilities, training a more general action flow model on internet-scale data and fine-tuning it in the robotic manipulation domain could be beneficial. Additionally, obtaining action flow from human videos could provide the model with a broader understanding of world knowledge.

\section{Simulation and tasks}
\subsection{Simulation and Dataset}
\begin{figure*}[htbp]
    \centering
    \includegraphics[width=\textwidth]{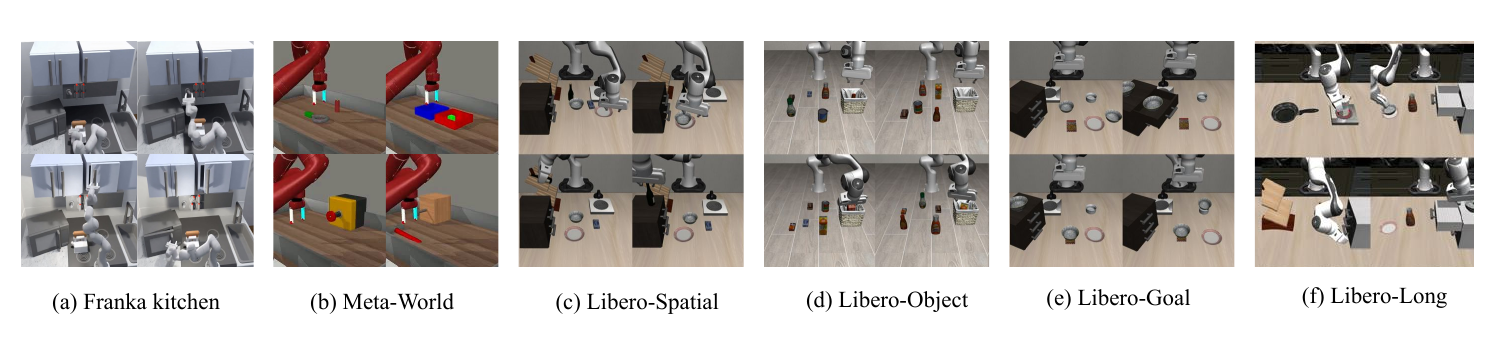}
    \caption{Various robot simulation environments. From left to right, there are (a) Franka Kitchen~\cite{fu2020d4rl} and (b) Meta-World~\cite{yu2020meta} for single manipulation task, and (c) Libero-Spatial~\cite{liu2024libero}, (d) Libero-Object~\cite{liu2024libero}, (e) Libero-Goal~\cite{liu2024libero} as well as (f) Libero-Long~\cite{liu2024libero} for long horizon manipulation task.}
    \label{fig:env}
\end{figure*}

We evaluated our method on single-task and multi-stage long-horizon manipulation tasks, simulations are shown in Figure\ref{fig:env}:
\begin{itemize}[leftmargin=1em]
    \item \textbf{Single Manipulation Task.} For Meta-World, we select five tasks: \texttt{assembly-v2} (putting a nut on a peg), \texttt{bin-picking-v2} (picking and placing a cube), \texttt{botton-press-v2} (horizontal push button switch), \texttt{drawer-close-v2} (closing a drawer), \texttt{hammer-v2} (hammering a nail). For Franka Kitchen, we select five tasks: \texttt{K-Slide} (open slide cabinet), \texttt{K-Hinge} (open the hinge cabinet), \texttt{K-Burner} (turn burner), \texttt{K-Light} (flick light switch), and \texttt{K-Microwave} (open microwave).
    \item \textbf{Long Horizon Manipulation Task.} We evaluated our approach on LIBERO, including Libero-Spatial, Libero-Object, Libero-Goal, and Libero-Long benchmarks. Each suite contains 10 tasks, and each instruction in these benchmarks typically contains 2-4 subtasks. Libero-Long is the most challenging, as it has a long horizon and complex manipulation tasks.
\end{itemize}
\label{sec:method}

\section{Impact of segmentation effect}
\begin{figure}[htbp]
    \centering
    \includegraphics[width=\columnwidth]{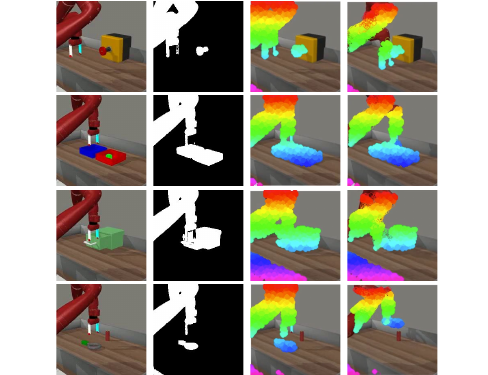}  
    % \caption{\textbf{Object masks with segmentation models.} Successful and failed object masks extracted by grounding-dino\cite{ren2024grounding} and segment-anything2\cite{ravi2024sam}.}
    \caption{\textbf{Object masks with segmentation models and corresponding optical flow.} Successful object masks extracted by grounding-dino\cite{ren2024grounding} and segment-anything2\cite{ravi2024sam}. And the changes in action flow in future frames. }
    \label{fig:seg}
\end{figure}

We try to leverage existing open vocalbulary segmentation methods to extract object masks required for action flow. Specifically, We use Language Segment-Anything based on GroundingDINO\cite{ren2024grounding} and Segment-Anything\cite{ravi2024sam}. Experiments were conducted in the Meta-World\cite{yu2020meta}, Franka Kitchen\cite{fu2020d4rl}, and Libero\cite{liu2024libero} environments, using predicted object masks instead of ground truth (GT) object masks. In the Meta-World setting, the segmentation success rate for small objects can reach 100\%. However, in the Libero environment, where object semantics are more complex and there are multiple objects, it is prone to failure when segmenting the target operational objects. Nonetheless, the robotic arm is always successfully segmented. In Franka kitchen, the objects manipulated are too large (such as cabinets, etc.), so we only segment the robotic arm to obtain the GT action flow. Qualitative object segmentation results are shown in Figure~\ref{fig:seg}.

% In Figure S1, we present a set of examples illustrating the generation of masks and the changes in action flow in future frames. 

\section{Implementation details}
% \subsection{Network and Training Details}
We train on 50 no-action-labeled videos for each task during the pretraining phase, consistent with baseline methods. After pretraining the action flow model, we train Dynamic Action Flow Integrator with 10 simulated demonstrations. 

We use a fixed pretrained CLIP-Text encoder\cite{radford2021learning} to encode textual descriptions. After encoding the textual descriptions with the encoder, we aggregate the output of the CLIP-Text encoder into a single vector and add it, along with image patches, to the time embeddings of the diffusion model. We use CoTracker3\cite{karaev2024cotracker3} and the segmentation model mask to obtain generated ground-truth trajectories to train the action flow generation. The checkpoint with the lowest validation loss is saved as our final model for the Dynamic Action Flow Integrator.

The hyperparameters related to Action Flow Diffusion are shown in Table~\ref{tab:hyp_flow}, while Table~\ref{tab:hyp_inte} lists the hyperparameters related to the Dynamic Action Flow Integrator. Table~\ref{tab:hyp_match} lists the hyperparameters related to the Matcher and reward function. These parameters are fixed and used for all experiments on the LIBERO benchmark dataset.

\begin{table}[ht]
    \centering
    \caption{\textbf{Hyperparameters of Action Flow Diffusion.}}
    \label{tab:hyp_flow}
    \begin{tabular}{lc}
        \hline
        Hyperparameters & Value \\
        \hline
        epoch & 100 \\
        batch size & 384 \\
        learning rate & 1e-4 \\
        dropout & 0\\
        optimizer & AdamW \\
        weight decay & 1e-4 \\
        lr scheduler & cosine \\
        lr warm up & 5 \\
        training objective & predict\_f\\
        loss function & l2 \\
        number of points & 256 \\
        action flow length & 16 \\
        action flow patch size & 4 \\
        \hline
    \end{tabular}
\end{table}

\begin{table}[ht]
    \centering
    \caption{\textbf{Hyperparameters of Dynamic Action Flow Integrator.}}
    \label{tab:hyp_inte}
    \begin{tabular}{lc}
        \hline
        Hyperparameters & Value \\
        \hline
        epoch & 100 \\
        batch size & 256 \\
        learning rate & 5e-4 \\
        optimizer & AdamW \\
        weight decay & 1e-4 \\
        N & 1\\
        lr scheduler & cosine \\
        lr warm up & 0 \\
        loss function & l2 \\
        augmentation & ColorJitter,RandomShift\\
        \hline
    \end{tabular}
\end{table}

\begin{table}[ht]
    \centering
    \caption{\textbf{Hyperparameters of Matcher.}}
    \label{tab:hyp_match}
    \begin{tabular}{lc}
        \hline
        Hyperparameters & Value \\
        \hline
        $\lambda_1$ & 0.3 \\
        $\lambda_2$ & 0.1 \\
        $\lambda_3$ & 0.6 \\
        $\alpha$ & 0.5 \\
        $\beta$ & 0.5 \\
        $\tau_{\text{total}}$ & 0.75\\
        \hline
    \end{tabular}
\end{table}

\section{More Experimental Results}
\subsection{Comparison of Different Policy Representations}
\begin{figure}[htbp]
    \centering
    \includegraphics[width=0.8\textwidth]{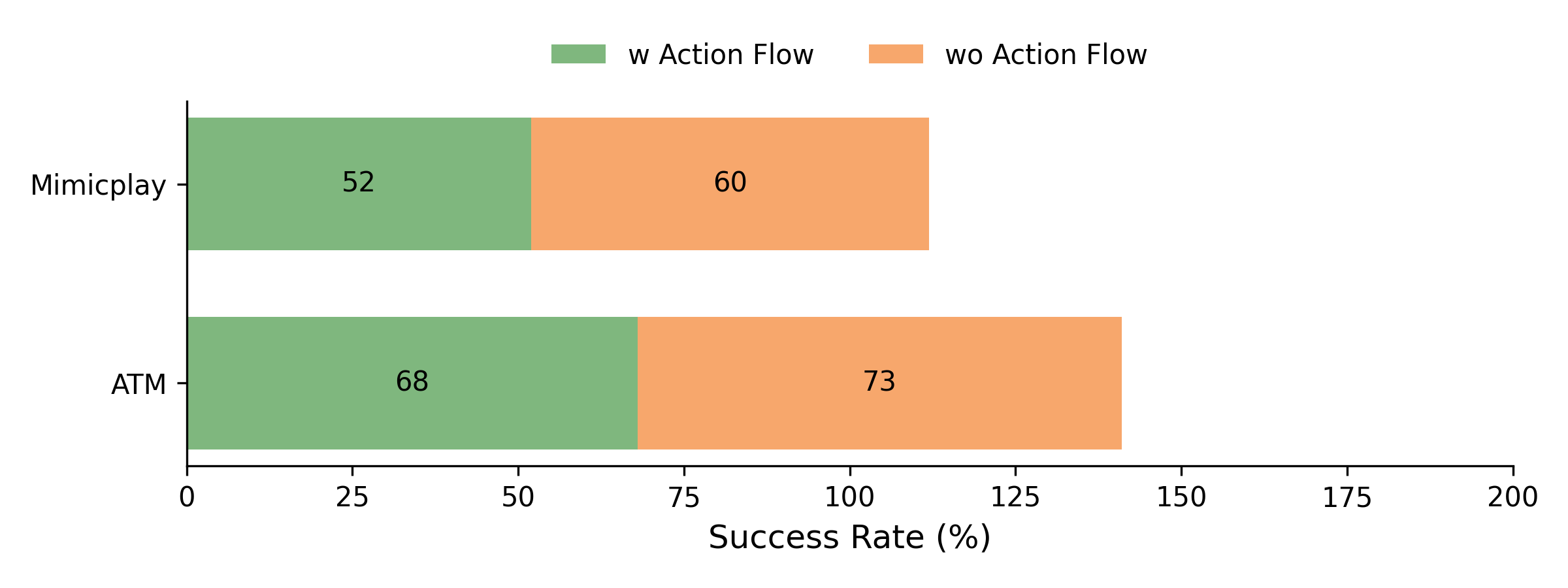}  
    \caption{\textbf{The baseline is replaced with Action Flow representation in the Meta-World simulator.} The average success rate of training policies with and without Action Flow is evaluated in 5 single-task settings.}
    \label{fig:baseline_flow}
\end{figure}

To validate the effectiveness of action flow, we replaced other representation methods with action flow, including MimicPlay's latent high-level plan\cite{wang2023mimicplay} and ATM's background-related any-point tracking\cite{wen2023any}. We did not modify the original model architecture or settings. As shown in Figure~\ref{fig:baseline_flow}, the replacement of representations improved the performance of the original methods. This indicates that the original observational data may contain a significant amount of task-irrelevant information, and some of the extracted latent representations lack the high correlation with robot actions and the dynamic information induced solely by actions that action flow provides. The improved results, as shown in the Figure~\ref{fig:baseline_flow}, further support this conclusion.
\subsection{Effect of Action Integrator.}
\begin{figure}[htbp]
    \centering
    \includegraphics[width=0.8\textwidth]{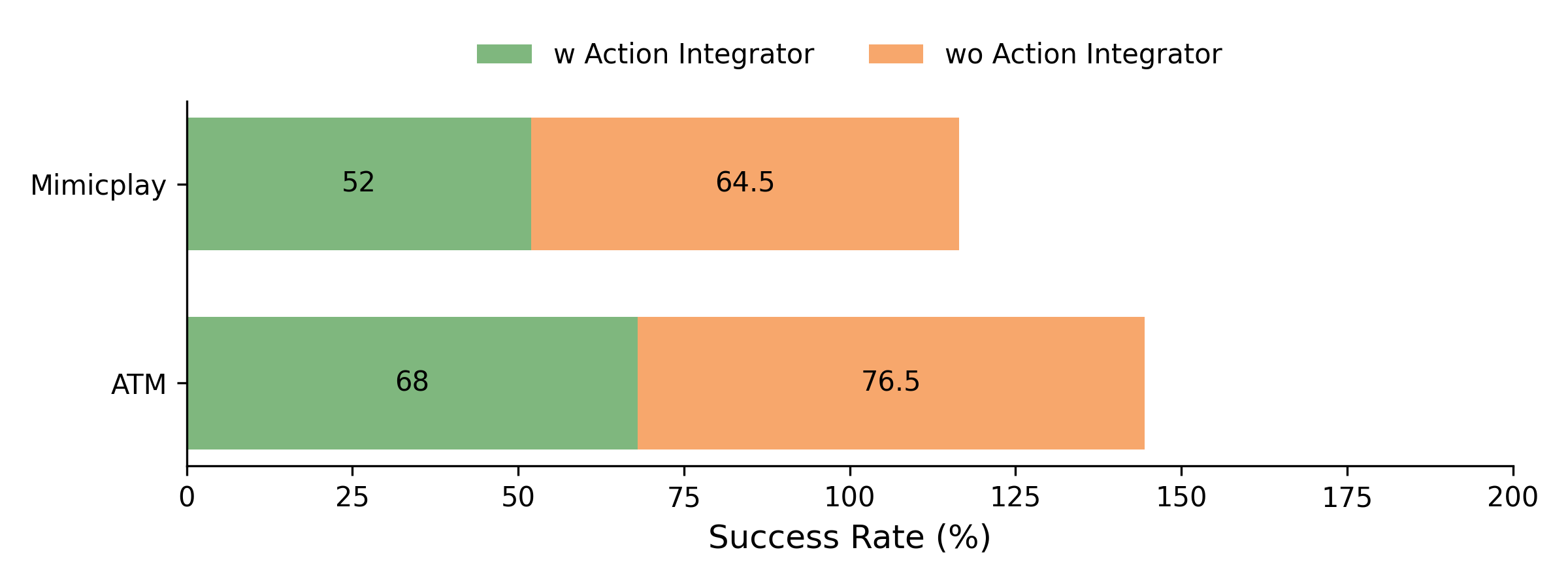}  
    \caption{\textbf{The baseline is added with Action Integrator in the Meta-World simulator.} The average success rate of training policies with and without Action Integrator is evaluated in 5 single-task settings.}
    \label{fig:baseline_integrator}
\end{figure}

Similarly, we integrated our action integration module into the baseline, enabling the baseline to store and access the working memory pool. To focus solely on the effect of action integration, we did not use action flow but instead matching with textual descriptions and images. The original model architecture and settings of the baseline remained unchanged. As shown in Figure ~\ref{fig:baseline_integrator}, this suggests that end-to-end direct estimation inevitably introduces deployment errors, and collecting more training data may not fundamentally solve this issue. In contrast, combining estimation with historical fusion for corrections can improve the accuracy of low-level control in a few-shot setting.

\section{Visualizations of action flow prediction and action prediction.}

\begin{figure*}[htbp]
    \centering     \includegraphics[width=1\textwidth]{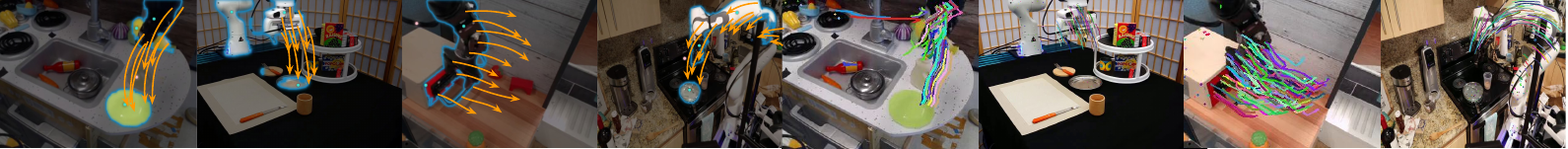}
    \caption{Real-world action flow experiments: ours (right 4) vs. manual annotations (left 4).}
    \label{fig:realflow}
\end{figure*}
% \vspace{-1.6em}

\begin{figure*}[h]
    \centering
    \includegraphics[width=\textwidth]{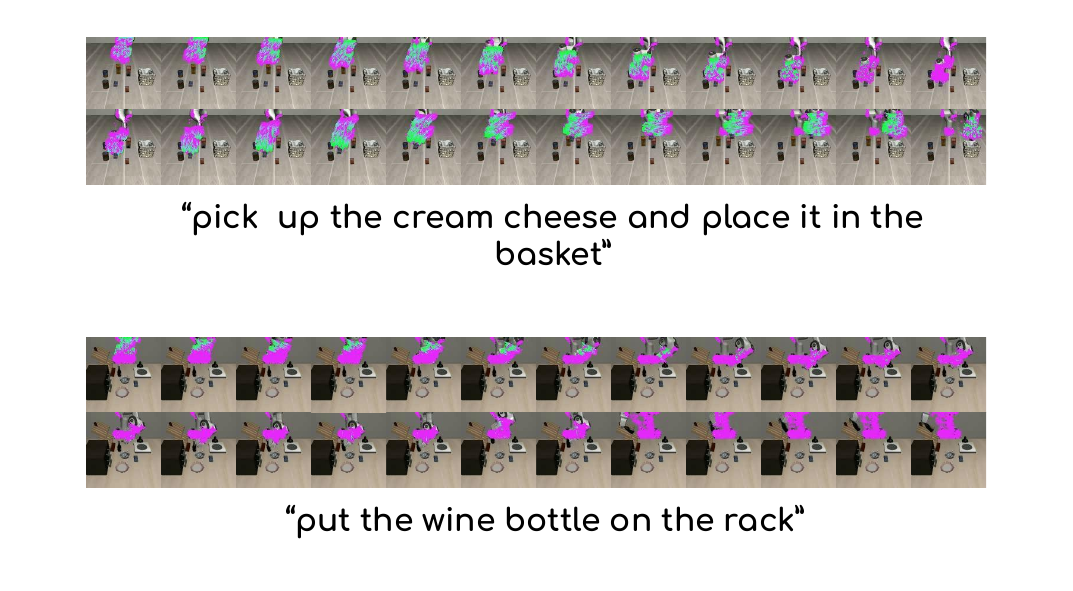} 
    \caption{\textbf{Action flow prediction results on Metaworld, Franka-Kitchen, and LIBERO.} From left to right, from top to bottom is the progress of the task. The pixels are initialized by random sampling of the segmented target motion region. The gradient from green to purple line indicates the direction of the action flow. The language instructions are placed below the image.}
    \label{fig:vis2}
\end{figure*}

We study the performance of action flow prediction. The qualitative results are shown in Figure~\ref{fig:vis2}. It can be observed that, given a task instruction and the segmentation predictions of the robotic arm and interactive objects in the image frames, our action flow model can predict the future trajectories of sparse pixel points in this region under the task condition. This eliminates background observations or object motions caused by non-robotic actions. 

After training the model on the video dataset, the predicted action flow can serve as a better conditional representation compared to images and text, providing effective representation and guidance for the subsequent Dynamic Action Flow Integrator. Despite initial segmentation failures of some target objects (e.g., the wine bottle) due to the presence of too many objects in the scene, ActionSink correctly predicts the future action flow of the robot. 

This predictive capability strongly positions action flow prediction as a potential module for future prediction and action integration, further demonstrating its applicability to real-world robotic tasks.
\begin{figure*}[h]
    \centering
    \includegraphics[width=\textwidth]{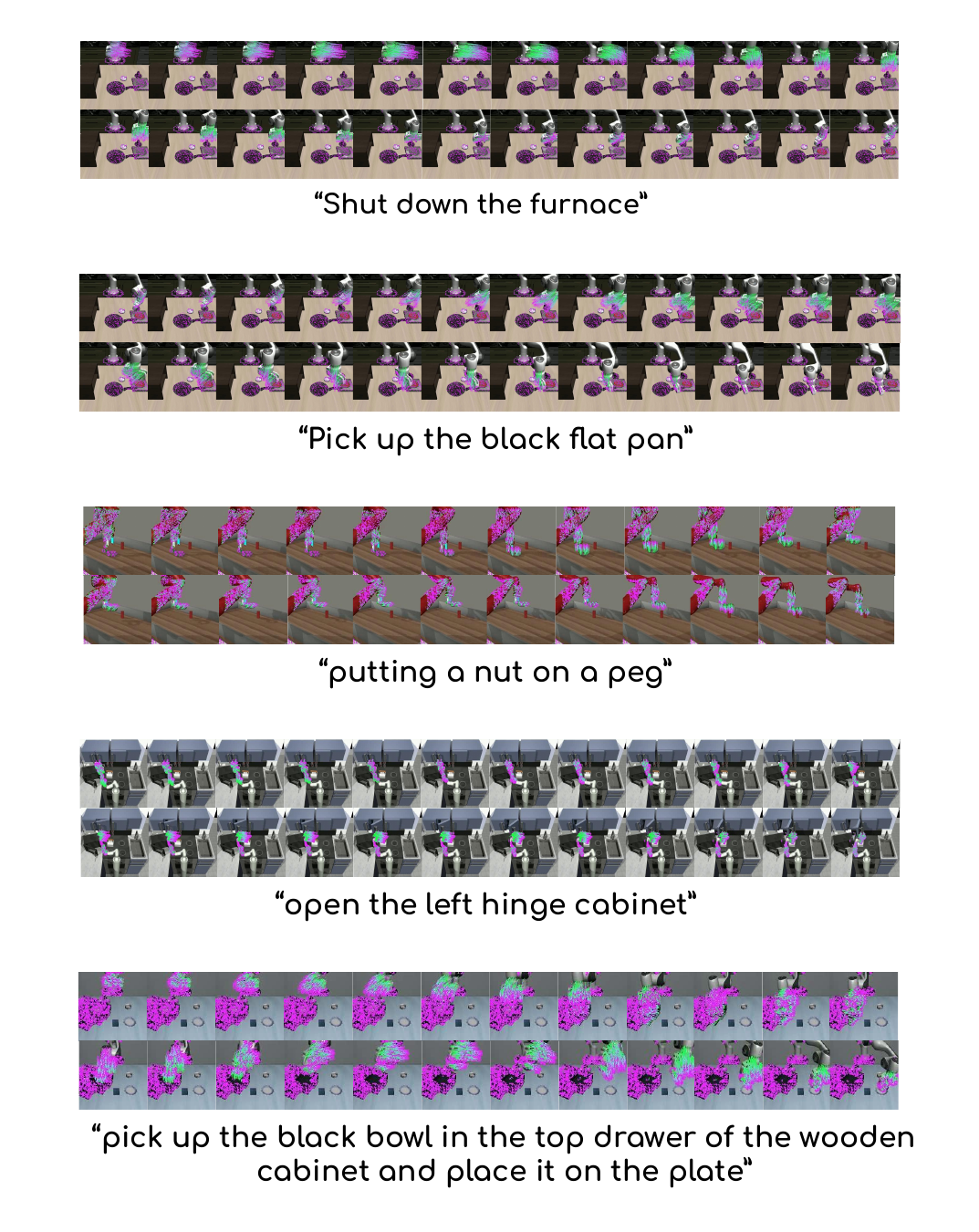} 
    \label{fig:vis1}
    % \caption{Action flow prediction results on Metaworld, Franka-Kitchen, and LIBERO. From left to right, from top to bottom is the progress of the task. The pixels are initialized by random sampling of the segmented target motion region. The gradient from green to purple line indicates the direction of the action flow. The language instructions are placed below the image.}
\end{figure*}

\end{document}